\pdfoutput=1
\documentclass[11pt]{article}

\usepackage[final]{ACL2023}

\usepackage{times}
\usepackage{latexsym}

\usepackage[T1]{fontenc}
\usepackage[utf8]{inputenc}

\usepackage{microtype}

\usepackage{inconsolata}
\usepackage{todonotes}

\usepackage{amsmath}
\usepackage{multirow}           % tabular cells spanning multiple
\usepackage{cleveref}
\usepackage{adjustbox}
\usepackage{booktabs}
\usepackage{cleveref}
\usepackage{graphicx}
\usepackage{subcaption}
\usepackage{comment}

\newcommand{\eg}{\textit{e.g\@.}}

\newcommand{\etal}{\textit{et~al\@.}}

\title{Revisiting Automated Prompting: Are We Actually Doing Better?}

\author{\textbf{ Yulin Zhou$^1$}
        \textbf{ Yiren Zhao$^2$}
        \textbf{ Ilia Shumailov$^3$}
        \textbf{ Robert Mullins$^1$}
        \textbf{ Yarin Gal$^3$} \\
        $^1$University of Cambridge \hspace{1em} $^2$Imperial College London \hspace{1em} $^3$University of Oxford \\
        \texttt{\href{mailto:yz709@cam.ac.uk}{yz709@cam.ac.uk}} \hspace{1em}
        \texttt{\href{mailto:a.zhao@imperial.ac.uk}{a.zhao@imperial.ac.uk}} \hspace{1em}
        \texttt{\href{mailto:ilia.shumailov@chch.ox.ac.uk}{ilia.shumailov@chch.ox.ac.uk}} \\
         \texttt{\href{mailto:robert.mullins@cl.cam.ac.uk}{robert.mullins@cl.cam.ac.uk}} \hspace{1em}
        \texttt{\href{mailto:yarin@cs.ox.ac.uk}{yarin@cs.ox.ac.uk}}
  }

\begin{document}
\maketitle

\begin{abstract}
Current literature demonstrates that Large Language Models (LLMs) are great few-shot learners, and \textit{prompting} significantly increases their performance on a range of downstream tasks in a few-shot learning setting. An attempt to automate human-led prompting followed, with some progress achieved. In particular, subsequent work demonstrates that automation can outperform fine-tuning in certain $K$-shot learning scenarios \cite{shin2020autoprompt, zhang2021differentiable}. In this paper, we revisit techniques for automated prompting on six different downstream tasks and a larger range of $K$-shot learning settings. We find that \emph{automated prompting does not consistently outperform simple manual prompting}. Our work suggests that, in addition to fine-tuning, \emph{manual prompting should be used as a baseline} in this line of research.
\end{abstract}
\section{Introduction}
Transformer-based Large Language Models (LLMs) are now considered foundation models for downstream tasks \cite{bommasani2021opportunities}. The \emph{pre-train then fine-tune} approach achieved state-of-the-art performance on a range of Natural Language Processing (NLP) tasks \cite{liu2019roberta,raffel2020exploring,brown2020language}. Unfortunately, in many NLP applications, the lack of high-quality labelled training data is a barrier to producing a model with good performance in the pre-train and then fine-tune approach. To address this issue, \emph{prompt-based learning} \cite{petroni2019language,schick2020exploiting,schick2020s,liu2021makes} emerged as a new paradigm for tuning a high-quality, pre-trained LLM in a \emph{few-shot learning} scenario, where only a few samples are available for downstream task learning.

\begin{figure}[t!]
    \centering
    \includegraphics[width=\hsize]{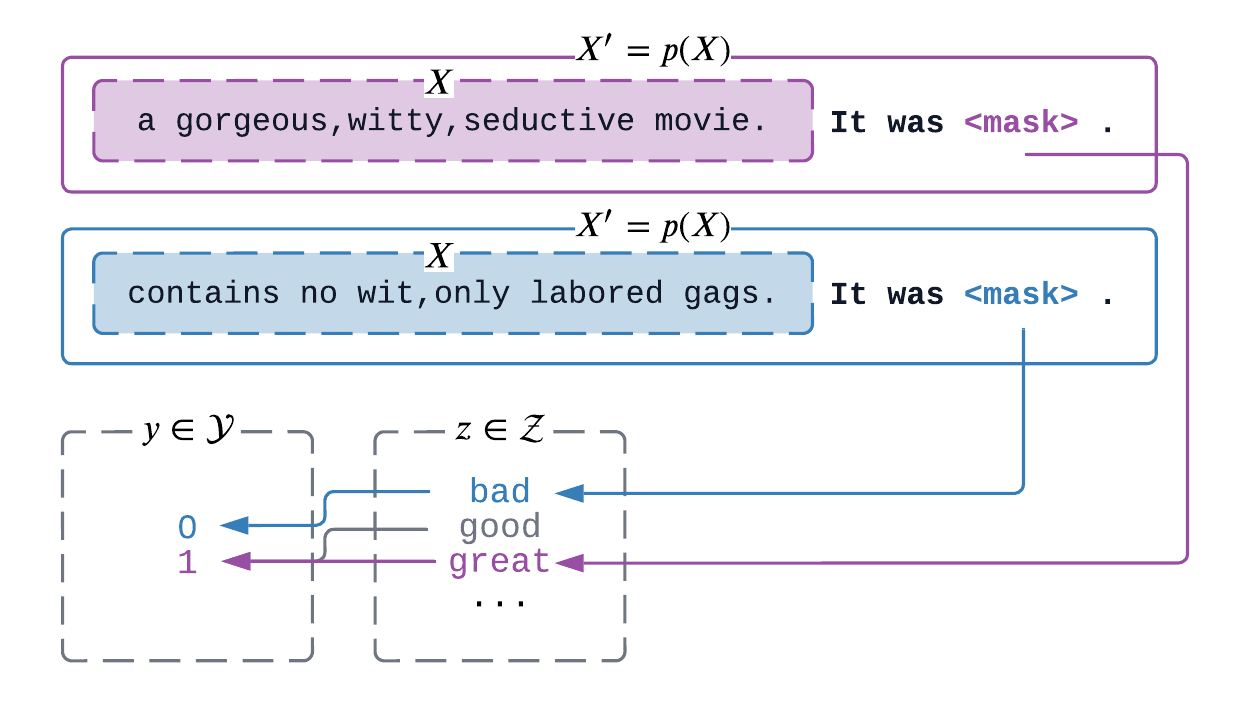}
\vspace{-15pt}

    \caption{Sentiment analysis with the prompt-based learning paradigm. Input $X'$ is the prompted input, and there is a many-to-one mapping between answers $z \in \mathcal{Z}$ and labels $y \in \mathcal{Y}$.}
    \label{fig:sst2_manual_prompt_example}
\vspace{-15pt}
\end{figure}

In the prompt-based learning paradigm (\Cref{fig:sst2_manual_prompt_example}), an input $X$ is modified using a template function $p$, also known as a prompting function and has one or more placeholders called mask tokens \textit{<mask>}, resulting in a prompted input $X' = p(X)$ \cite{liu2021pre}. Additionally, a verbaliser designs an answer domain $\mathcal{Z}$, so that for an output label domain $\mathcal{Y}$, there is a many-to-one mapping for an answer $z \in \mathcal{V}_y \subseteq \mathcal{Z}$ to an output label $y \in \mathcal{Y}$ in accordance with the downstream task.
Considering a language model $f_{o}$ pre-trained on a large corpus of text, such as Wikipedia, the goal of prompt-based learning is to fine-tune it on a small dataset of prompted inputs $X'$ and corresponding output $y$, in order to produce a high-quality language model $f_{p}$ capable of generating an answer $z$ for a given input $X$.

Prompting formulates downstream tasks such as sentiment analysis and text classification to cloze completion (also known as filling in the blanks). Furthermore, using prompts and fine-tuning allows models to gain superior few-shot learning capabilities \cite{lester2021power,schick2020exploiting,shin2020autoprompt}. 
Despite the relative success of prompt-based learning, the design of prompts can be a challenging task.
As a result, many research studies sought to \emph{automate} the process of designing suitable prompts for downstream tasks \cite{liu2021p,zhang2021differentiable,shin2020autoprompt}. The motivation for automating prompt design is usually two-fold: first, manually designing prompts can be time-consuming; and second, automated ones can often provide better performance. In this work, we question \emph{the second motivation} and demonstrate that \emph{existing automated prompts do not consistently outperform their manual counterparts} under various $K$-shot learning setups.
In this paper, we make the following contributions:

\begin{itemize}
    \item  We thoroughly investigate automated prompts and demonstrate that they do not consistently outperform manual prompts, even when the latter are created using basic heuristics and selected among a small number of options (\Cref{sec:eval:res}).
    \item We show empirically that fine-tuning only serves a strong baseline when $K \geq 100$ in a $K$-shot learning setup (\Cref{sec:eval:res}).
    \item By visualising the prompts generated by auto-prompting, we explain why these prompts are not necessarily better than manually designed ones (\Cref{sec:discussion:visual}). 
    \item 
    Supported by our empirical evidence and evaluation, we strongly recommend that \emph{future research should consider manual prompts as a simple yet effective baseline}.
\end{itemize}
\section{Related Work}
The rise of the \emph{prompting-based learning paradigm} comes with the development of LLMs \cite{brown2020language}, which were demonstrated to be good few-shot learners \cite{liu2021gpt}.
To begin with, researchers focused on manually crafted prompts for downstream tasks \cite{petroni2019language,liu2021pre,scao2021many,zhao2021calibrate, schick2020exploiting}, yet soon shifted towards automated prompt designs. Schick \etal~investigated how to automatically identify label words for a prompt \cite{schick2020exploiting, schick2020s}, while Shin~\etal~proposed AutoPrompt, a framework for automatically generating prompts for various tasks, through a gradient-based search \cite{shin2020autoprompt}. Gao~\etal~used another LLM, T5 \cite{raffel2020exploring}, to generate both the prompting templates and verbaliser answer domains~\cite{gao2020making}. Han~\etal~incorporated logic rules into prompt designs, combining several simple sub-prompts according to these rules \cite{han2022ptr}. All of the above mentioned methods are based on the assumption that the prompt design has to rely on discrete tokens.

Liu \etal~and Lester \etal~demonstrated that prompts could be trainable continuous embeddings, or soft prompts, instead of discrete tokens. These soft prompts can be learned with a frozen language model (LLM) on a target task \cite{liu2021gpt,lester2021power,zhang2021differentiable}. Liu \etal~further discovered that Deep Prompts, which are soft prompts used in every layer of the model, allow for scaling to large LLMs for complex natural language processing (NLP) tasks \cite{liu2021p}. Zhang \etal~developed Differentiable Prompts, which put the label tokens design of the prompt into a continuous space and optimised it jointly with soft prompts \cite{zhang2021differentiable}. An extensive evaluation was conducted by Zhang \etal~on various downstream tasks.

Most of the work on automating prompt design mentioned above has two major motivations: to reduce the amount of time it takes to design prompts manually; and to potentially gain better performance, since manual prompt formats can be sub-optimal \cite{zhang2021differentiable}. While the first motivation may be valid in some cases, it largely depends on the  task complexity and the amount of data available -- it is sometimes possible for non-experts to design a prompt sufficient for simple tasks with a large amount of data. The principal focus of this work, however, is on the second motivation: \emph{can automated prompts really outperform manual prompts in a consistent manner?}
A comparison between automated and manual prompts is lacking in current research. To our knowledge, automated prompting methods focus solely on comparing to fine-tuning in a few-shot learning setup, while a comparisons to manual prompting methods remain unexplored. 
In this paper, we consider AutoPrompt (Auto) \cite{shin2020autoprompt} and Differential Prompt (Diff) \cite{zhang2021differentiable} as representatives, where one is based on discrete tokens, while the other is based on continuous embeddings. 
We compare them with manually designed prompts and fine-tuning without prompting on various tasks. 
\section{Evaluation}
\subsection{Experiment setup}
A robust framework was developed to assess prompting model performance under $K$-shot learning scenarios where only $K$ samples per class are available for the training and validation datasets. Three prompting models were re-implemented: LM-BFF (manual) \cite{gao2020making}, AutoPrompt (Auto) \cite{shin2020autoprompt}, and DART (Diff) \cite{zhang2021differentiable} models. During prompt-based learning, each prompting model is allowed to fine-tune the parameters of the pre-trained language model using the limited training and validation datasets.
\subsubsection{Datasets and Model} We conducted comprehensive experiments on six datasets to compare the performance of prompting models fine-tuned on the pre-trained RoBERTa-large model \cite{liu2019roberta}. \Cref{tab:dataset_setup} in \Cref{sec:appendix:dataset_setup} shows we picked three sentiment analysis and three textural entailment tasks. 
% \newpage
% \noindent
\subsubsection{Prompt Templates and Verbalisers}
\label{sec:eval:template}
We design prompts to concatenate the input text and the \textit{<mask>} token, alongside a verbaliser that maps from the answer domain to the output label domain.
Manually designed prompts and verbalisers are adapted from the Public Pool of Prompts \cite{ppp2022} and previous work on prompting \cite{gao2020making, attack2022}. For each dataset, we selected four to six prompt-and-verbaliser pairs, compared their performance under the same $K = 16$ few-shot scenario, and picked the best-performing pair for further experiments with different $K$ values. Detailed manually designed prompts and verbalisers, as well as their performance measures, are illustrated in \Cref{tab:manual_select}, and the best-performing pairs are summarised in \Cref{tab:manual_prompts} in \Cref{sec:appendix:prompt_manual}. 

An automated discrete prompt replaces the template with trigger tokens \textit{<T>}. Following the same settings used in AutoPrompt \cite{shin2020autoprompt}, we inserted ten trigger tokens between the input text and the \textit{<mask>} token. Under a $K$-shot scenario, the verbaliser mapping is automatically generated from the train and validation dataset, each with $K$ samples per class. Table \ref{tab:auto_prompts} in \Cref{sec:appendix:prompt_design} shows the automated discrete prompts and verbalisers for each dataset. 
A differential prompt starts from the manually designed prompt but treats both the template and the verbaliser as a collection of differentiable parameters.

Take the dataset SST2 as an example: a suitable manually designed prompt could be \texttt{``<sentence> . It was <mask> .''} with a verbaliser $\{\texttt{bad} \mapsto 0, \texttt{good} \mapsto {1}\}$; An automated discrete prompt could be \texttt{``<sentence> <T> ... <T> <mask> .''} with ten trigger tokens \texttt{<T>}. 

% \noindent 
\subsubsection{Hyper-parameters}
We conducted a beam search using the AdamW optimiser \cite{adamw2017} for the optimal batch size, learning rate and weight decay for each set of experiments with the same dataset and $K$-shot value. Each experiment is run with $100$ epochs and an early stopping value of $5$, \textit{i.e.}, when the validation loss is non-decreasing for $5$ epochs. The detailed hyper-parameters used in each set of experiments are listed in~\Cref{tab:hyper_param}, and details on the evaluation metrics are in~\Cref{sec:appendix:hyper}. 

\subsection{Main Results}
\label{sec:eval:res}
\Cref{tab:glue} illustrates the performance of various prompting strategies. We observe that manual prompts exhibit the best performance in 13 out of the 24 setups (6 different datasets and 4 different $K$s), and the second-best performance in 8 of them. Automated prompts (both Auto and Diff) only show a clear advantage in TWEETS-HATE-OFFENSIVE when $K=100$. The baseline in \Cref{tab:glue} is direct fine-tuning on the $K$ samples.

We also see that automated prompts can be catastrophically ineffective in certain setups. For example, as shown in \Cref{tab:auto_prompts}, Auto performs much worse than Manual or Baseline in MNLI-MATCHED when $K=100$. Diff also significantly underperforms Manual in TWEETS-HATE-OFFENSIVE when $K=16$. In later parts of this section, we provide an analysis of the generated prompts and explore the reasons for this phenomenon. 
Finally, we demonstrate that Baseline sometimes performs well when $K$ is large. This is seen in SST2 when $K=100, 1000$ and also ENRON-SPAM when $K=100$. % Summarise important points at the end
In general, we make the following observations:
\begin{itemize}
	\item Manual prompting outperforms automated prompting (Auto and Diff) with different $K$-shot setups on most tasks.
	\item Automated prompting sometimes cannot even outperform fine-tuning, \eg~MNLI-MISMATCHED $K=100, 1000$.
	\item When $K$ is small, prompting can greatly improve performance, \eg~on SST2 and MNLI.
	\item Automated prompting can fail catastrophically (\eg~ MNLI-MISMATCHED  $K=1000$) and have a high variance in performance (\eg~$15.5$ standard deviation on SST2), while manual prompting is more robust.
\end{itemize}
\begin{table*}[!ht]
\centering
\adjustbox{max width=\textwidth}{
	\begin{tabular}{c | llll | llll }
	\toprule
	\multicolumn{1}{c}{ }                      
	& \multicolumn{4}{c}{SST2}                      
	& \multicolumn{4}{c}{QNLI} \\
	$K$ 
        & Baseline & Auto	& Diff	& Manual 
	& Baseline & Auto	& Diff	& Manual \\
	\midrule
        $8$   
	% SST2
	& $59.8 \pm 8.6$        
	& $51.7 \pm 1.9$           
	& $\boldsymbol{88.0 \pm 1.6}$
	& $\underline{77.6 \pm 4.6}$           

	% QNLI 
	& $49.9 \pm 1.0$        
	& $\underline{51.5 \pm 0.7}$           
	& $50.5 \pm 2.1$             
	& $\boldsymbol{54.6 \pm 2.8}$           

	\\
	$16$   
	% SST2
	& $72.1 \pm 15.0$        
	& $70.1 \pm 3.9$           
	& $\boldsymbol{87.8 \pm 0.7}$
	& $\underline{86.9 \pm 1.6}$           

	% QNLI 
	& $49.9 \pm 0.2$        
	& $53.4 \pm 1.3$           
	& $\underline{59.5 \pm 3.6}$             
	& $\boldsymbol{74.1 \pm 1.2}$           

	\\
	$100$  
 	% SST2
	& $\boldsymbol{89.6 \pm 0.5}$             
	& $83.5 \pm 4.3$              
	& $88.6 \pm 0.7$            
	& $\underline{89.4 \pm 1.0}$       
        % QNLI
        & $78.9 \pm 2.3$
        & $74.0 \pm 4.3$
        & $\underline{80.2 \pm 2.1}$
        & $\boldsymbol{82.7 \pm 0.7}$
        \\
	$1000$
        % SST2
	& $\boldsymbol{92.7 \pm 0.2}$           
	& $\underline{92.5 \pm 0.2}$
        & $90.1 \pm 0.7$
        & $92.3 \pm 0.2$
        % QNLI
        & $\underline{87.2 \pm 1.0}$
        & $83.2 \pm 3.8$
        & $85.2 \pm 1.1$
        & $\boldsymbol{88.0 \pm 0.3}$
        \\
	\midrule
	\multicolumn{1}{c}{}                      
	& \multicolumn{4}{c}{MNLI-Matched}                      
	& \multicolumn{4}{c}{MNLI-Mismatched} \\
	$K$
	& Baseline & Auto	& Diff	& Manual  
	& Baseline & Auto	& Diff	& Manual  
	\\
	\midrule
        $8$
        % matched
        & $34.6 \pm 2.4$
        & $34.2 \pm 1.1$
        & $\underline{51.3 \pm 1.1}$
        & $\boldsymbol{55.7 \pm 3.3}$
        % mismatched
        & $33.8 \pm 0.8$
        & $33.8 \pm 0.5$
        & $\underline{47.6 \pm 3.0}$
        & $\boldsymbol{56.0 \pm 1.4}$ \\
        
        $16$
        % matched
        & $33.3 \pm 0.2$
        & $34.9 \pm 0.7$
        & $\boldsymbol{61.4 \pm 1.5}$
        & $\underline{60.2 \pm 3.7}$
        % mismatched
        & $32.8 \pm 1.3$
        & $35.6 \pm 0.8$
        & $\underline{59.4 \pm 1.1}$
        & $\boldsymbol{60.2 \pm 2.7}$ \\
        
        $100$
        % matched
        & $63.1 \pm 1.3$
        & $42.3 \pm 0.5$
        & $\underline{72.1 \pm 0.8}$
        & $\boldsymbol{74.1 \pm 1.2}$
        % mismatched
        & $\underline{73.6 \pm 2.1}$
        & $39.5 \pm 1.0$
        & $73.3 \pm 1.2$
        & $\boldsymbol{77.0 \pm 1.2}$ \\
        
        $1000$
        % matched
        & $\underline{82.7 \pm 0.5}$
        & $72.9 \pm 2.3$
        & $80.0 \pm 0.8$
        & $\boldsymbol{83.2 \pm 0.3}$
        % mismatched
        & $\underline{84.3 \pm 0.5}$
        & $76.6 \pm 3.7$
        & $82.0 \pm 0.4$
        & $\boldsymbol{85.0 \pm 0.2}$ \\
        \midrule
	\multicolumn{1}{c}{}                      
	& \multicolumn{4}{c}{ENRON-SPAM}                      
	& \multicolumn{4}{c}{TWEETS-HATE-OFFENSIVE} \\
	$K$
	& Baseline & Auto	& Diff	& Manual  
	& Baseline & Auto	& Diff	& Manual  
	\\
        \midrule
        $8$
        % enron-spam
        & $49.1 \pm 36.6$
        & $\underline{73.4 \pm 6.0}$
        & $\boldsymbol{80.7 \pm 5.7}$
        & $67.9 \pm 12.2$
        % tweets
        & $14.5 \pm 9.5$
        & $12.1 \pm 4.6$
        & $\boldsymbol{32.5 \pm 7.1}$
        & $\underline{25.8 \pm 16.5}$ \\
        $16$
        % enron-spam
        & $84.2 \pm 4.0$
        & $80.5 \pm 2.6$
        & $\underline{88.0 \pm 2.3}$
        & $\boldsymbol{89.4 \pm 3.0}$
        % tweets
        & $\underline{38.0 \pm 4.1}$
        & $42.5 \pm 2.6$
        & $37.2 \pm 7.7$
        & $\boldsymbol{46.7 \pm 2.5}$ \\
        $100$
        % enron-spam
        & $\boldsymbol{97.1 \pm 0.4}$
        & $90.8 \pm 0.4$
        & $\underline{96.3 \pm 0.8}$
        & $\underline{96.3 \pm 0.5}$
        % tweets
        & $44.9 \pm 0.9$
        & $\underline{51.4 \pm 3.4}$
        & $\boldsymbol{59.7 \pm 2.8}$
        & $47.0 \pm 0.8$ \\	
        $1000$
        % enron-spam
        & $98.0 \pm 0.5$
        & $97.0 \pm 0.7$
        & $\boldsymbol{99.0 \pm 0.1}$
        & $\underline{98.7 \pm 0.2}$
        % tweets
        & $66.5 \pm 1.5$
        & $66.8 \pm 1.8$
        & $\boldsymbol{67.7 \pm 3.3}$
        & $\underline{67.5 \pm 2.1}$ \\
        \bottomrule
        \end{tabular}
 }
 \caption{The performance of various prompting methods on RoBERTa-large \cite{liu2019roberta} was assessed using numbers reported as percentages, with a mean and standard deviation across five independent runs. The best and second-best performing methods are represented in bold and underlined fonts, respectively. The baseline is fine-tuning only without any prompting, while Auto, Diff, and Manual correspond to AutoPrompt \cite{shin2020autoprompt}, Differential Prompt \cite{zhang2021differentiable}, and LM-BFF \cite{gao2020making}, respectively.}
 \label{tab:glue}
\end{table*}

\begin{figure*}[!ht]
\begin{subfigure}{.33\textwidth}
  \centering
  \includegraphics[width=\linewidth]{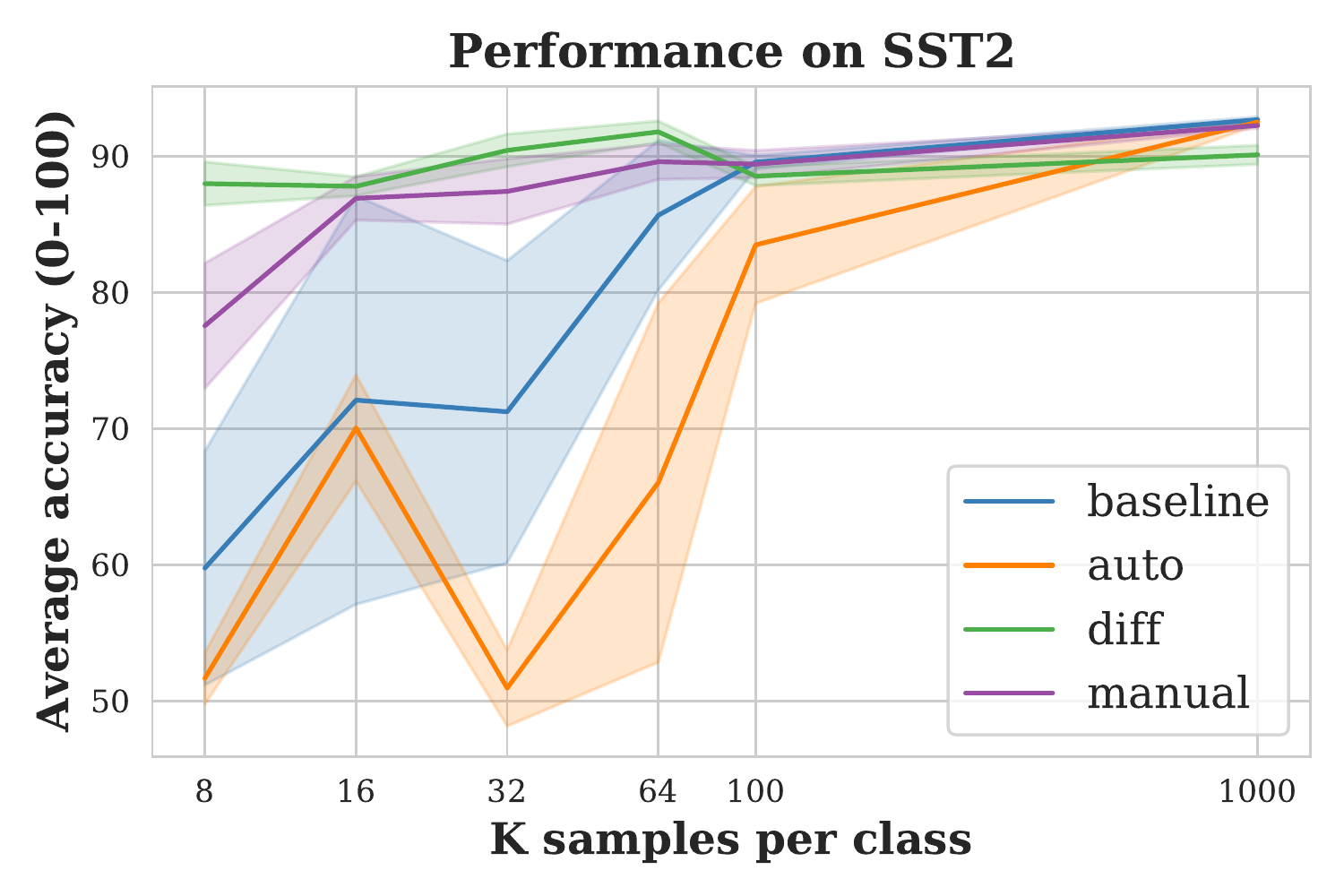}
  \caption{SST2}
  \label{fig:sst}
\end{subfigure}%
\begin{subfigure}{.33\textwidth}
  \centering
  \includegraphics[width=\linewidth]{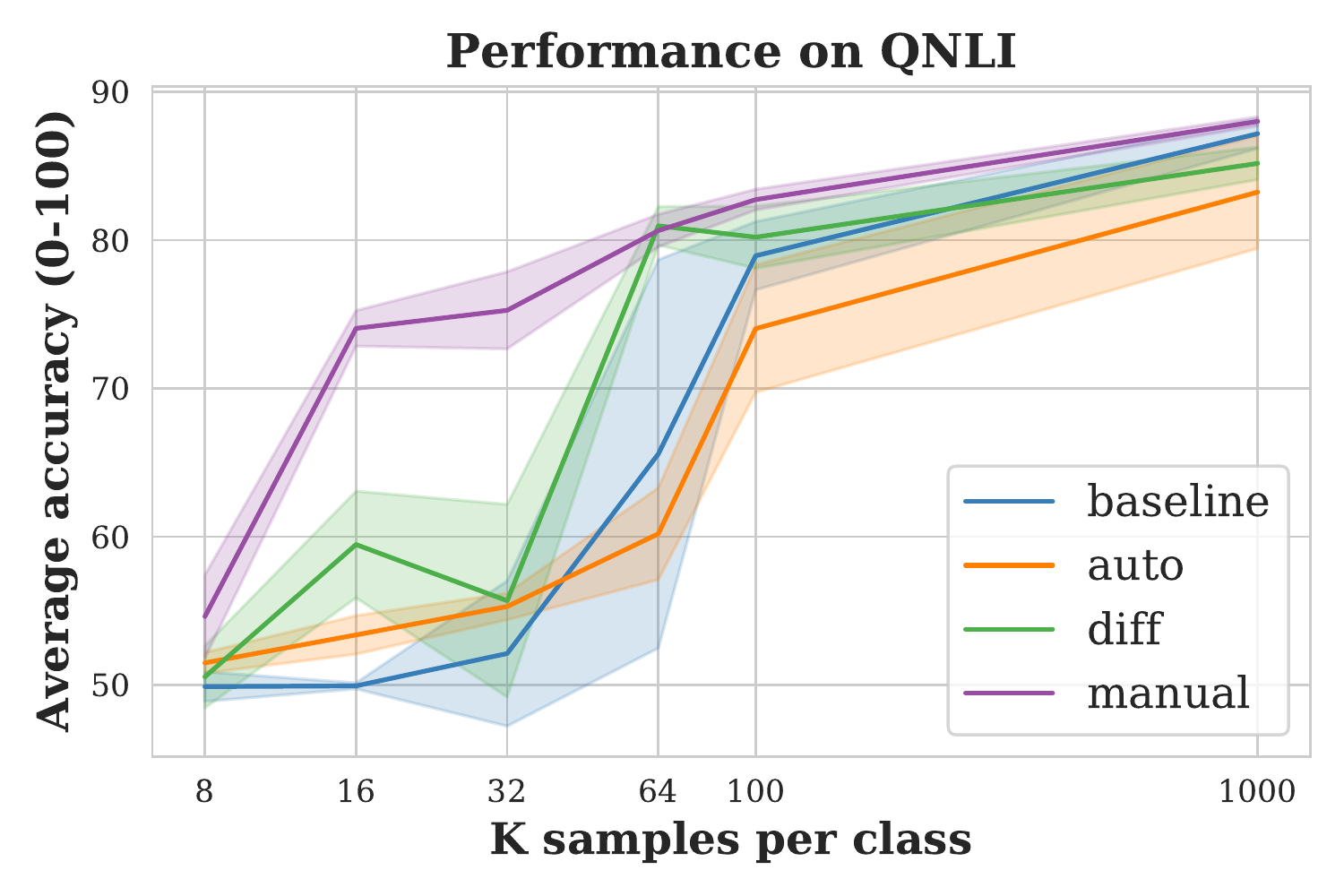}
  \caption{QNLI}
  \label{fig:qnli}
\end{subfigure}
\begin{subfigure}{.33\textwidth}
  \centering
  \includegraphics[width=\linewidth]{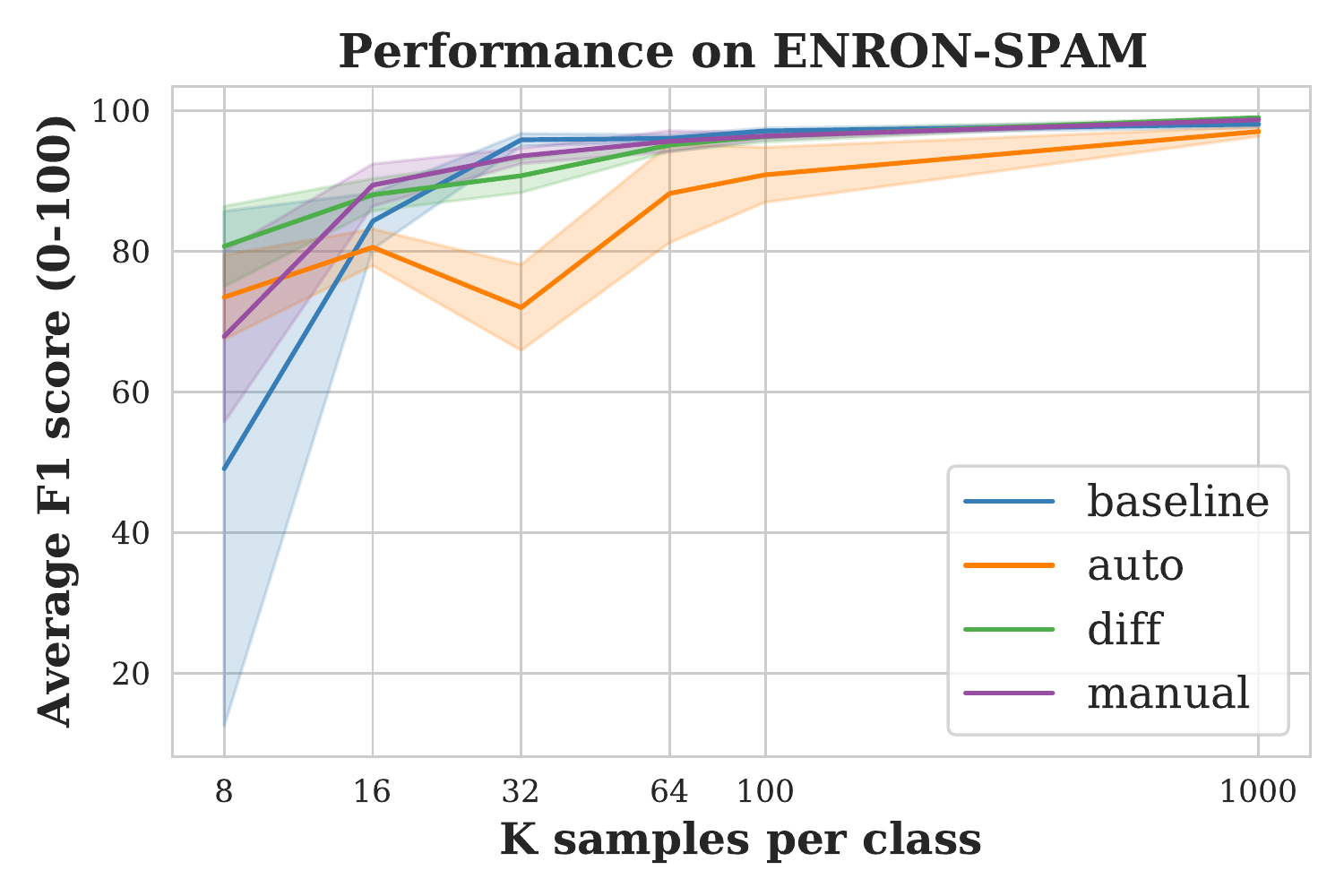}
  \caption{ENRON-SPAM}
  \label{fig:enron}
\end{subfigure}
\vspace{-10pt}
\caption{The performance of prompting models on the datasets SST2, QNLI \cite{gluedataset2018} and ENRON-SPAM \cite{enronspam2006} is shown for a wider range of $K$ values. The solid line plots the mean accuracy across five independent runs, and is bounded by one standard deviation on both sides.}
\label{fig:more_k}
\vspace{-10pt}
\end{figure*}

% \textbf{Discussion}
\subsection{More $K$-shot Experiments}
\label{sec:eval:kshot}
% Show on one or two tasks, how a wider range of $K$-shot values affects performance in different prompting setups.
\Cref{fig:more_k} demonstrates the performance of different prompting styles with more $K$ values on SST2, QNLI \cite{gluedataset2018} and ENRON-SPAM \cite{enronspam2006}. 

We observe that the performance of all methods starts to converge with larger $K$ values, which is consistent with existing literature \cite{shin2020autoprompt}. 
It is also worth mentioning that the automated prompting methods do not consistently outperform manual prompting on this large range of $K$ values. More results are available in \Cref{sec:appendix:kshot}.

% \begin{figure}[t]
%     \centering
%     \includegraphics[width=\hsize]{./images/SST2_prompting_performance.pdf}
%     \caption{The performance of prompting models on the dataset SST2 \cite{gluedataset2018} for a wider range of $K$ values. The solid line plots the mean accuracy across five independent runs, and is bounded by one standard deviation on both sides.}
%     \label{fig:sst2_wider_k}
% \end{figure}

% \begin{figure}[t]
%     \centering
%     \includegraphics[width=\hsize]{./images/QNLI_prompting_performance.pdf}
%     \caption{The performance of prompting models on the dataset QNLI \cite{gluedataset2018} for a wider range of $K$ values. The solid line plots the mean accuracy across five independent runs, and is bounded by one standard deviation on both sides.}
%     \label{fig:qnli_wider_k}
% \end{figure}

\subsection{Visualizing Auto-prompts}
\label{sec:discussion:visual}
As previously discussed, automated prompting can sometimes fail catastrophically. \Cref{tab:auto_prompts} summarises all the automated discrete prompts and verbaliser answer domains. Since the answer domain is generated from the $K$ samples per class, it may not be general enough or optimal for the entire dataset. 
On the other hand, manual prompts and verbalisers are designed based on common knowledge that humans possess from countless examples encountered in daily life. One possible improvement idea on AutoPrompt is to start with a manually designed prompt and update both the prompt and the verbaliser through a gradient-based search in an iterative manner.

% \noindent
\subsection{Limitations}
All prompting methods are trying to extract knowledge from the Large Language Models (LLMs). Our paper compares their knowledge extraction abilities. Thus, the performance of RoBERTa-large can serve as a reference point and provide insights for other LLMs. However, it is still necessary to assess each large language model independently to understand its capabilities comprehensively.

We only tested a handful of simple manual prompt-and-verbaliser pairs which are included in \Cref{tab:manual_select,tab:manual_prompts}. It is entirely possible that there is a lot of room for improvement in the design of manual prompt-and verbaliser pairs, thus providing us a even stronger baseline.
We have opted to use ten trigger tokens in Auto, in alignment with the experiment settings originally presented in the AutoPrompt paper \cite{shin2020autoprompt}. However, since the verbaliser  domains generated under few-shot learning settings are noisy, reducing the number of trigger tokens may improve performance.
\section{Conclusion}
In this paper, we revisit the results generated from automated prompting, and show that \emph{automated prompting cannot consistently outperform simple manual prompting on a variety of tasks}. We also demonstrate that the performance of automated prompting is heavily dependent on the amount of data available, and in some cases can even be worse than fine-tuning. On the other hand, manual prompting is more robust to the amount of data available, and can have similar performance to fine-tuning if not outperforming. We take a closer look at the prompts and verbalisers generated by automated discrete prompting (AutoPrompt) and point out that few-shot learning settings make it challenging to generate prompts and verbalisers that perform well. We hope that this work will motivate researchers to use manual prompts as a general baseline.

\section*{Acknowledgment}
The authors would like to thank the anonymous reviewers for their helpful suggestions.

\bibliography{anthology,custom}
\bibliographystyle{acl_natbib}

\appendix

\label{sec:appendix}
\section{Model and infrastructure details}
\label{sec:appendix:model_setup}
The RoBERTa-large model \cite{liu2019roberta} is pre-trained on a large corpus of raw English text using masked language modelling (MLM) objective; it contains $354$ million parameters.

All our experiments are run parallelly on 4 NVIDIA Tesla V100 GPUs; for smaller $K$ values (e.g., $K = 100$), most experiments require less than 1 GPU hour, while a setting with a larger $K$ value (e.g., $K = 1000$) may require 2 GPU hours.

\section{Dataset details}
\label{sec:appendix:dataset_setup}
We conducted comprehensive experiments on six datasets (SST2, QNLI, MNLI-MATCHED, MNLI-MISMATCHED, ENRON-SPAM and TWEETS-HATE-OFFENSIVE) to compare the performance of prompting models fine-tuned on the pre-trained RoBERTa-large model. As shown in \Cref{tab:dataset_setup}, we picked three sentiment analysis and three textural entailment tasks. 
Among the six, three are binary classifications (SST2, QNLI and ENRON-SPAM), while the remaining datasets have three categories each (MNLI-MATCHED, MNLI-MISMATCHED and TWEETS-HATE-OFFENSIVE). 

\begin{table*}[!ht]
\centering
\adjustbox{max width=\hsize}{
	\begin{tabular}{c | c | c | p{9cm} }
	\toprule
	\textbf{Dataset} & \# \textbf{Class} & \textbf{Test Sample} & \textbf{Description} \\
	\midrule
        % SST2
	\multirow{3}{*}{SST2} 
        & \multirow{3}{*}{2} & \multirow{3}{*}{33674}
        & A sentiment analysis task on movie reviews from the GLUE benchmark \cite{gluedataset2018}. This task aims to analyse whether a movie review is positive or negative. \\
        \midrule
        
        % QNLI
	\multirow{4}{*}{QNLI} 
        & \multirow{4}{*}{2} & \multirow{4}{*}{5463}
        & A textual entailment task on question-answer pairs from the GLUE benchmark \cite{gluedataset2018}. The objective is to determine whether the context sentence contains the answer to the question. \\

        \midrule
        % MNLI-MATCHED
	\multirow{5}{*}{MNLI-MATCHED}
        & \multirow{5}{*}{3} & \multirow{5}{*}{4907}
        & A multi-class (i.e., entailment, neutral, contradiction) textual entailment task on premise-hypothesis pairs from the GLUE benchmark \cite{gluedataset2018}. Matched version only preserves pairs within the same genre (e.g., science fiction, speech). \\

        \midrule
        % MNLI-MISMATCHED
	\multirow{4}{*}{MNLI-MISMATCHED}
        & \multirow{4}{*}{3} & \multirow{4}{*}{4916}
        &  Same as MNLI-MATCHED, the mismatched version is a textual entailment task on premise-hypothesis pairs from the GLUE benchmark \cite{gluedataset2018}, but it only preserves pairs within different genres.\\

        \midrule
        % ENRON-SPAM
	\multirow{2}{*}{ENRON-SPAM} 
        & \multirow{2}{*}{2} & \multirow{2}{*}{15858}
        &  A safety critical binary sentiment analysis task determining whether an email text is a spam \cite{enronspam2006}.\\

        \midrule    
        % TWEETS-HATE-OFFENSIVE
	\multirow{3}{*}{TWEETS-HATE-OFFENSIVE} 
        & \multirow{3}{*}{3} & \multirow{3}{*}{12391}
        &  A safety critical multi-class sentiment analysis task which aims to classify whether a tweet text contains hate speech, offensive speech or neither \cite{tweetsho2017}. \\
	\toprule
        \end{tabular}
 }
 \caption{Six datasets selected in the project. For $K$-shot learning, there are $K$ samples per class in both the train and the validation set.}
 \label{tab:dataset_setup}
\end{table*}

\section{Manual prompt-and-verbaliser designs}
\begin{table*}[!ht]
\centering
\adjustbox{max width=\hsize}{
	\begin{tabular}{l | c | c | l | c | c}
	\toprule                      
	\multicolumn{3}{c}{SST2}                      
	& \multicolumn{3}{c}{QNLI} \\
	\textbf{Prompt Design} & \textbf{Answer $\mapsto$ Label} & \textbf{Accuracy}
    & \textbf{Prompt Design} & \textbf{Answer $\mapsto$ Label} & \textbf{Accuracy}\\
	\midrule  
	% SST2
	\multirow{6}{*}{\texttt{<sentence> . It was <mask> .}}
    & {\texttt{terrible} $\mapsto$ 0, \texttt{great} $\mapsto$ 1}
    & $86.0 \pm 2.7$
	% QNLI 
	& \texttt{<question> ? <mask> , <sentence> .}      
	& \multirow{5}{*}{\{\texttt{Yes} $\mapsto$ 0,}
    & $64.5 \pm 4.8$
	\\ 
    % SST2
    & {\texttt{bad} $\mapsto$ 0, \texttt{good} $\mapsto$ 1}
    & $\boldsymbol{86.9 \pm 1.6}$ 
	% QNLI 
	& \texttt{<question> . <mask> , <sentence> .}      
	& \multirow{5}{*}{ \texttt{No} $\mapsto$ 1\}}
    & $60.5 \pm 2.3$
    \\
    % SST2
    & {\texttt{dog} $\mapsto$ 0, \texttt{cat} $\mapsto$ 1}
    & $84.7 \pm 3.4$
	% QNLI 
	& \texttt{<question> ? <mask> <sentence> .}   
	&  
    & $68.7 \pm 3.2$
	\\
    % SST2
    & {\texttt{cat} $\mapsto$ 0, \texttt{dog} $\mapsto$ 1}
    & $68.7 \pm 6.9$
	% QNLI 
	& \texttt{<sentence> ? <mask> , <question> .}       
	&  
    & $\boldsymbol{74.1 \pm 1.2}$
    \\
    % SST2
    & {\texttt{great} $\mapsto$ 0, \texttt{terrible} $\mapsto$ 1}  
    & $67.4 \pm 5.0$
	% QNLI 
	& \texttt{<question> <mask> <sentence>}       
	&  
    & $50.0 \pm 0.2$
    \\
    % SST2 
    &   
    & 
	% QNLI 
	& \texttt{<sentence> ? <mask> , <question>}       
	&  
    & $66.7 \pm 10.2$ \\
    \midrule
	\multicolumn{3}{c}{MNLI-Matched}               
	& \multicolumn{3}{c}{MNLI-Mismatched} \\
	\textbf{Prompt Design} & \textbf{Answer $\mapsto$ Label} & \textbf{Accuracy}
    & \textbf{Prompt Design} & \textbf{Answer $\mapsto$ Label} & \textbf{Accuracy}
	\\
	\midrule
        % matched
        \texttt{<premise> ? <mask> , <hypothesis> .}
        & \multirow{4}{*}{\{\texttt{Yes} $\mapsto$ 0,}
        & $\boldsymbol{60.2 \pm 3.7}$
        % mismatched
        & \texttt{<premise> ? <mask> , <hypothesis> .}
        & \multirow{4}{*}{\{\texttt{Yes} $\mapsto$ 0,}
        & $\boldsymbol{60.2 \pm 2.7}$
        \\
        % matched
        \texttt{<premise> . <mask> , <hypothesis> .}
        & \multirow{4}{*}{\texttt{Maybe} $\mapsto$ 1,}
        & $58.6 \pm 4.8$
        % mismatched
        & \texttt{<premise> . <mask> , <hypothesis> .}
        & \multirow{4}{*}{\texttt{Maybe} $\mapsto$ 1,}
        & $56.3 \pm 1.5$
        \\	
        % matched
        \texttt{<premise> ? <mask> <hypothesis> .}
        & \multirow{4}{*}{\texttt{No} $\mapsto$ 2\}}
        & $55.6 \pm 1.7$
        % mismatched
        & \texttt{<premise> ? <mask> <hypothesis> .}
        & \multirow{4}{*}{ \texttt{No} $\mapsto$ 2\}}
        & $58.4 \pm 1.1$
        \\
        % matched
        \texttt{<hypothesis> ? <mask> , <premise> .}
        & 
        &  $51.9 \pm 4.2$
        % mismatched
        & \texttt{<hypothesis> ? <mask> , <premise> .}
        & 
        & $57.9 \pm 0.8$
        \\
        % matched
        \texttt{<premise> <mask> <hypothesis>}
        & 
        & $51.2 \pm 4.2$
        % mismatched
        & \texttt{<premise> <mask> <hypothesis>}
        & 
        & $49.4 \pm 2.4$
        \\
        % matched
        \texttt{<hypothesis> ? <mask> , <premise>}
        &
        & $52.4 \pm 2.9$
        % mismatched
        & \texttt{<hypothesis> ? <mask> , <premise>}
        & 
        & $56.0\pm 1.0$ \\
        \midrule                   
	\multicolumn{3}{c}{ENRON-SPAM}                      
	& \multicolumn{3}{c}{TWEETS-HATE-OFFENSIVE} \\
	\textbf{Prompt Design} & \textbf{Answer $\mapsto$ Label} & \textbf{F1 score}
    & \textbf{Prompt Design} & \textbf{Answer $\mapsto$ Label} & \textbf{F1 score}
	\\
        \midrule
        % enron-spam
        \texttt{<mask> : <text> .}
        & {\texttt{ham} $\mapsto$ 0, \texttt{spam} $\mapsto$ 1}
        & $82.8 \pm 1.9$
        % tweets
        & \texttt{<tweet> . This post is <mask> .}
        & \multirow{2}{*}{\{\texttt{hateful} $\mapsto$ 0,}
        & $\boldsymbol{46.7 \pm 2.5}$
        \\
        % enron-spam
        \texttt{This is a <mask> : <text> .}
        & {\texttt{ham} $\mapsto$ 0, \texttt{spam} $\mapsto$ 1}
        & $82.8 \pm 2.8$
        % tweets
        & \texttt{This post is <mask> : <tweet> .}
        & \multirow{2}{*}{\texttt{ offensive} $\mapsto$ 1,}
        & $40.3 \pm 3.8$ \\	
        % enron-spam
        \texttt{<mask> email : <text> .}
        & {\texttt{genuine} $\mapsto$ 0, \texttt{spam} $\mapsto$ 1}
        & $\boldsymbol{89.4 \pm 3.0}$
        % tweets
        & \texttt{<tweet> . This was <mask> .}
        & \multirow{2}{*}{\texttt{ harmless} $\mapsto$ 2\}}
        & $39.8 \pm 4.5$ \\
        % enron-spam
        \texttt{<text> . This was a <mask> .}
        & {\texttt{ham} $\mapsto$ 0, \texttt{spam} $\mapsto$ 1}
        & $76.8 \pm 3.3$
        % tweets
        & \texttt{<mask> speech : <tweet> .}
        & 
        & $36.8 \pm 11.7$ \\
        \toprule
    \end{tabular}
 }
 \caption{The prompt-and-verbaliser pairs are tested under the few-shot scenario $K = 16$, and the best-performing pair is highlighted in bold. The mean and standard deviation of scores are computed across five independent runs.}
 \label{tab:manual_select}
\end{table*}

\label{sec:appendix:prompt_manual}
In the \textbf{Prompt Templates and Verbalisers} part in \Cref{sec:eval:template}, we discussed how we picked the best-performing prompt-and-verbaliser pairs. We show the picked manual prompt with their picked verbalisers in \Cref{tab:manual_select}, covering SST2, QNLI, MNLI-MATCHED, MNLI-MISMATCHED, ENRON-SPAM and TWEETS-HATE-OFFENSIVE.
The underlying mechanism for finding a good manual prompt is detailed in \Cref{sec:eval:template}. As one can see in these tables, the manual prompts used are very simple and requires minimal domain knowledge.

\section{Generated Auto-prompts}
\label{sec:appendix:prompt_design}
% Manual prompts
\begin{table*}[!ht]
\centering
\adjustbox{max width=\hsize}{
	\begin{tabular}{c | l | l }
	\toprule
	\textbf{Dataset} & \textbf{Prompt Design} & \textbf{Answer $\mapsto$ Label} \\
	\midrule
        % SST2
	SST2 
        & \texttt{<sentence> . It was <mask> .}
        & {\texttt{bad} $\mapsto$ 0, \texttt{good} $\mapsto$ 1} \\

        % QNLI
	QNLI 
        & \texttt{<sentence> ? <mask> , <question> .}
        & {\texttt{Yes} $\mapsto$ 0, \texttt{No} $\mapsto$ 1} \\

        % MNLI-MATCHED
	MNLI-MATCHED 
        & \texttt{<premise> ? <mask> , <hypothesis> .}
        & {\texttt{Yes} $\mapsto$ 0, \texttt{Maybe} $\mapsto$ 1, \texttt{No} $\mapsto$ 2} \\

        % MNLI-MISMATCHED
	MNLI-MISMATCHED 
        & \texttt{<premise> ? <mask> , <hypothesis> .}
        & {\texttt{Yes} $\mapsto$ 0, \texttt{Maybe} $\mapsto$ 1, \text{No} $\mapsto$ 2} \\

        % ENRON-SPAM
	ENRON-SPAM 
        & \texttt{<mask> email : <text> .}
        & {\texttt{genuine} $\mapsto$ 0, \texttt{spam} $\mapsto$ 1} \\

        % TWEETS-HATE-OFFENSIVE
	TWEETS-HATE-OFFENSIVE 
        & \texttt{<tweet> . This post is <mask> .}
        & {\texttt{hateful} $\mapsto$ 0, \texttt{offensive} $\mapsto$ 1, \texttt{harmless} $\mapsto$ 2} \\
	
        \toprule
        \end{tabular}
 }
 \caption{Summarised for each dataset, the best-performing manual prompt and verbaliser.}
 \label{tab:manual_prompts}
\end{table*}

% Auto prompts
\begin{table*}[!ht]
\centering
\adjustbox{max width=\hsize}{
	\begin{tabular}{c | c | c | c }
	\toprule
	\textbf{Task} & \textbf{Prompt design} & $\boldsymbol{K}$ & \textbf{Answer $\boldsymbol{\mapsto}$ Label} \\
	\midrule
        % SST2
        \multirow{6}{*}{SST2}
        & 
        & $8$
        & {\texttt{impunity} $\mapsto$ 0, \texttt{ASHINGTON} $\mapsto$ 1} \\

        & \multirow{3}{*}{\texttt{<sentence> <T> <T> <T> <T> <T>}}
        & $16$
        & {\texttt{worthless} $\mapsto$ 0, \texttt{Kom} $\mapsto$ 1} \\

        & \multirow{3}{*}{\texttt{<T> <T> <T> <T> <T> <mask> .}}
        & $32$
        & {\texttt{Worse} $\mapsto$ 0, \texttt{å¤©} $\mapsto$ 1} \\
        
        &
        & $64$
        & {\texttt{horrible} $\mapsto$ 0, \texttt{magic} $\mapsto$ 1} \\

        & 
        & $100$
        & {\texttt{worse} $\mapsto$ 0, \texttt{å¤©} $\mapsto$ 1} \\

        & 
        & $1000$
        & {\texttt{worse} $\mapsto$ 0, \texttt{Excellent} $\mapsto$ 1} \\

        \midrule
        % QNLI
        \multirow{6}{*}{QNLI}
        &
        & $8$
        & {\texttt{implement} $\mapsto$ 0, \texttt{defensively} $\mapsto$ 1} \\

        & \multirow{3}{*}{\texttt{<question> <mask> <T> <T> <T> <T> <T>}} 
        & $16$
        & {\texttt{counter} $\mapsto$ 0, \texttt{Bits} $\mapsto$ 1} \\

        & \multirow{3}{*}{\texttt{<T> <T> <T> <T> <T> <sentence>}}
        & $32$
        & {\texttt{Meteor} $\mapsto$ 0, \texttt{univers} $\mapsto$ 1} \\
        
        &
        & $64$
        & {\texttt{ormon} $\mapsto$ 0, \texttt{stood} $\mapsto$ 1} \\

        & 
        & $100$
        & {\texttt{idelines} $\mapsto$ 0, \texttt{opard} $\mapsto$ 1} \\

        & 
        & $1000$
        & {\texttt{Ģ} $\mapsto$ 0, \texttt{overloaded} $\mapsto$ 1} \\

        \midrule
        % MNLI-MATCHED
        
        \multirow{6}{*}{MNLI-MATCHED}
        & 
        & $8$
        & {\texttt{efforts} $\mapsto$ 0, \texttt{democratically} $\mapsto$ 1, \texttt{Congratulations} $\mapsto$ 2} \\
        
        & \multirow{3}{*}{\texttt{<premise> <mask> <T> <T> <T> <T> <T>}}
        & $16$
        & {\texttt{OWN} $\mapsto$ 0, \texttt{hypocritical} $\mapsto$ 1, \texttt{examiner} $\mapsto$ 2} \\

        & \multirow{3}{*}{\texttt{<T> <T> <T> <T> <T> <hypothesis>}}
        & $32$
        & {\texttt{Alicia} $\mapsto$ 0, \texttt{historians} $\mapsto$ 1, \texttt{BF} $\mapsto$ 2} \\

        &
        & $64$
        & {\texttt{tweets} $\mapsto$ 0, \texttt{onboard} $\mapsto$ 1, \texttt{Anniversary} $\mapsto$ 2} \\

        & 
        & $100$
        & {\texttt{filmmakers} $\mapsto$ 0, \texttt{combat} $\mapsto$ 1, \texttt{absence} $\mapsto$ 2} \\

        & 
        & $1000$
        & {\texttt{thus} $\mapsto$ 0, \texttt{MED} $\mapsto$ 1, \texttt{independent} $\mapsto$ 2} \\

        \midrule
        % MNLI-MISMATCHED
        
        \multirow{6}{*}{MNLI-MISMATCHED}
        & 
        & $8$
        & {\texttt{Whilst} $\mapsto$ 0, \texttt{oka} $\mapsto$ 1, \texttt{smokers} $\mapsto$ 2} \\
        
        & \multirow{3}{*}{\texttt{<premise> <mask> <T> <T> <T> <T> <T>}}
        & $16$
        & {\texttt{Accordingly} $\mapsto$ 0, \texttt{)?} $\mapsto$ 1, \texttt{foreigners} $\mapsto$ 2} \\

        & \multirow{3}{*}{\texttt{<T> <T> <T> <T> <T> <hypothesis>}}
        & $32$
        & {\texttt{ibliography} $\mapsto$ 0, \texttt{qa} $\mapsto$ 1, \texttt{Governments} $\mapsto$ 2} \\

        &
        & $64$
        & {\texttt{LER} $\mapsto$ 0, \texttt{jack} $\mapsto$ 1, \texttt{foreigners} $\mapsto$ 2} \\

        & 
        & $100$
        & {\texttt{HEL} $\mapsto$ 0, \texttt{gaming} $\mapsto$ 1, \texttt{imperialism} $\mapsto$ 2} \\

        & 
        & $1000$
        & {\texttt{Vladimir} $\mapsto$ 0, \texttt{acting} $\mapsto$ 1, \texttt{dislike} $\mapsto$ 2} \\

        \midrule
        % ENRON-SPAM
        \multirow{6}{*}{ENRON-SPAM}
        &
        & $8$
        & {\texttt{Reviewer} $\mapsto$ 0, \texttt{Pure} $\mapsto$ 1} \\

        & \multirow{3}{*}{\texttt{ <question> <mask> <T> <T> <T> <T> <T>}} 
        & $16$
        & {\texttt{debian} $\mapsto$ 0, \texttt{Discount} $\mapsto$ 1} \\

        & \multirow{3}{*}{\texttt{<T> <T> <T> <T> <T> <sentence>}}
        & $32$
        & {\texttt{hillary} $\mapsto$ 0, \texttt{Vampire} $\mapsto$ 1} \\
        
        &
        & $64$
        & {\texttt{schedules} $\mapsto$ 0, \texttt{Romance} $\mapsto$ 1} \\

        & 
        & $100$
        & {\texttt{subcommittee} $\mapsto$ 0, \texttt{Beauty} $\mapsto$ 1} \\

        & 
        & $1000$
        & {\texttt{committee} $\mapsto$ 0, \texttt{ophobic} $\mapsto$ 1} \\

        \midrule
        % TWEETS-HATE-OFFENSIVE
        
        \multirow{6}{*}{TWEETS-HATE-OFFENSIVE}
        & 
        & $8$
        & {\texttt{Slater} $\mapsto$ 0, \texttt{herself} $\mapsto$ 1, \texttt{issued} $\mapsto$ 2} \\
        
        & \multirow{3}{*}{\texttt{<premise> <mask> <T> <T> <T> <T> <T>}}
        & $16$
        & {\texttt{kicking} $\mapsto$ 0, \texttt{her} $\mapsto$ 1, \texttt{selections} $\mapsto$ 2} \\

        & \multirow{3}{*}{\texttt{<T> <T> <T> <T> <T> <hypothesis>}}
        & $32$
        & {\texttt{athi} $\mapsto$ 0, \texttt{herself} $\mapsto$ 1, \texttt{vernight} $\mapsto$ 2} \\

        &
        & $64$
        & {\texttt{racist} $\mapsto$ 0, \texttt{Marie} $\mapsto$ 1, \texttt{skies} $\mapsto$ 2} \\

        & 
        & $100$
        & {\texttt{racist} $\mapsto$ 0, \texttt{vaginal} $\mapsto$ 1, \texttt{Miracle} $\mapsto$ 2} \\

        & 
        & $1000$
        & {\texttt{homophobia} $\mapsto$ 0, \texttt{b***h} $\mapsto$ 1, \texttt{heavens} $\mapsto$ 2} \\
	
        \bottomrule
        \end{tabular}
 }
 \caption{Auto prompts designed alongside with the automatically generated verbalisers for each dataset.}
 \label{tab:auto_prompts}
\end{table*}
In the \textbf{Prompt Templates and Verbalisers} part in \Cref{sec:eval:template}, we also mentioned that an automated discrete prompt replaces the template with trigger tokens \textit{<T>}. Following the same settings used in AutoPrompt \cite{shin2020autoprompt}, we inserted ten trigger tokens between the input text and the \textit{<mask>} token. All automated discrete prompts and their automatically generated verbalisers are listed in \Cref{tab:auto_prompts}. In contrast to the manual prompts shown in \Cref{sec:appendix:prompt_manual}, the auto-prompts generated are now more complex.

\begin{table*}[!ht]
\centering
\adjustbox{scale=.7}{
	\begin{tabular}{c | c | c c c | c | c | c c c}
	\toprule
	\textbf{Dataset} & \textbf{Model} & \textbf{Batch Size} & \textbf{$\boldsymbol{\eta}$} & \textbf{$\boldsymbol{w_d}$} & \textbf{Dataset} & \textbf{Model} & \textbf{Batch Size} & \textbf{$\boldsymbol{\eta}$} & \textbf{$\boldsymbol{w_d}$}   \\
	\midrule
        % SST2
        % Auto
        \multirow{3}{*}{SST2} 
        & Auto
        &  8
        & 1e-5
        & 0.01 
        
        % QNLI 
        % Auto
        & \multirow{3}{*}{QNLI} 
        & Auto
        &  4 
        & 2e-5
        & 0.1 \\

        % SST2
        % Diff 
        & Diff
        &  8
        & 1e-5
        & 0.01 
        % QNLI
        % Diff 
        &
        & Diff
        &  4
        & 1e-5
        & 0.1 \\

        % SST2
        % Manual
        & Manual
        &  4
        & 2e-5
        & 0.01 
        % QNLI
        % Manual
        &
        & Manual
        &  4
        & 2e-5
        & 0.01 \\

        \midrule
        % MNLI-MATCHED 
        % Auto
        \multirow{3}{*}{MNLI-MATCHED} 
        & Auto
        &  4 
        & 2e-5
        & 0.01 
        % MNLI-MISMATCHED 
        % Auto
        & \multirow{3}{*}{MNLI-MISMATCHED} 
        & Auto
        &  4 
        & 2e-5
        & 0.01 \\
        
        % MNLI-MATCHED
        % Diff 
        & Diff
        &  4
        & 1e-5
        & 0.01 
        % MNLI-MISMATCHED
        % Diff 
        &
        & Diff
        &  8
        & 1e-5
        & 0.01 \\

        % MNLI-MATCHED
        % Manual
        & Manual
        &  4
        & 2e-5
        & 0.01 
        % MNLI-MISMATCHED
        % Manual
        &
        & Manual
        &  4
        & 2e-5
        & 0.01 \\

        \midrule
        % ENRON-SPAM
        % Auto
        \multirow{3}{*}{ENRON-SPAM} 
        & Auto
        &  8 
        & 1e-5
        & 0.01 
        % TWEETS-HATE-OFFENSIVE 
        % Auto
        & \multirow{3}{*}{TWEETS-HATE-OFFENSIVE } 
        & Auto
        &  8 
        & 2e-5
        & 0.1 \\

        % ENRON-SPAM
        % Diff 
        & Diff
        &  8
        & 2e-5
        & 0.0
        % TWEETS-HATE-OFFENSIVE
        % Diff 
        &
        & Diff
        &  8
        & 2e-5
        & 0.0 \\

        % ENRON-SPAM
        % Manual
        & Manual
        &  8
        & 2e-5
        & 0.05

        % TWEETS-HATE-OFFENSIVE
        % Manual
        &
        & Manual
        &  8
        & 2e-5
        & 0.1 \\

        \toprule
        \end{tabular}
 }
 \caption{Details of the selected hyper-parameters, including batch size, learning rate $\eta$ and weight decay $w_d$ for each set of experiments with the same dataset and prompting model.}
 \label{tab:hyper_param}
\end{table*}
\section{Hyper-parameters and evaluation metrics for training}
\label{sec:appendix:hyper}
In terms of the evaluation metrics which measure the performance of the prompting models, we utilised two different metrics according to the nature of the datasets: (1) Multi-class classification accuracy for balanced datasets SST2, QNLI, MNLI-MATCHED and MNLI-MISMATCHED. (2) F1 score captures both precisions and recalls for safety-critical or unbalanced datasets ENRON-SPAM and TWEETS-HATE-OFFENSIVE.

\Cref{tab:hyper_param} provides details for the training setups. We show the batch sizes, learning rates and weight decay values used in the experiments. We also show the optimal hyper-parameters for each set of experiments with the same dataset and prompting model. For example, the optimal hyper-parameters for the dataset SST2 with the prompting model Auto are batch size $8$, learning rate $10^{-5}$ and weight decay $0.01$.

\begin{figure*}[!t]
\begin{subfigure}{.33\textwidth}
  \centering
    \includegraphics[width=\linewidth]{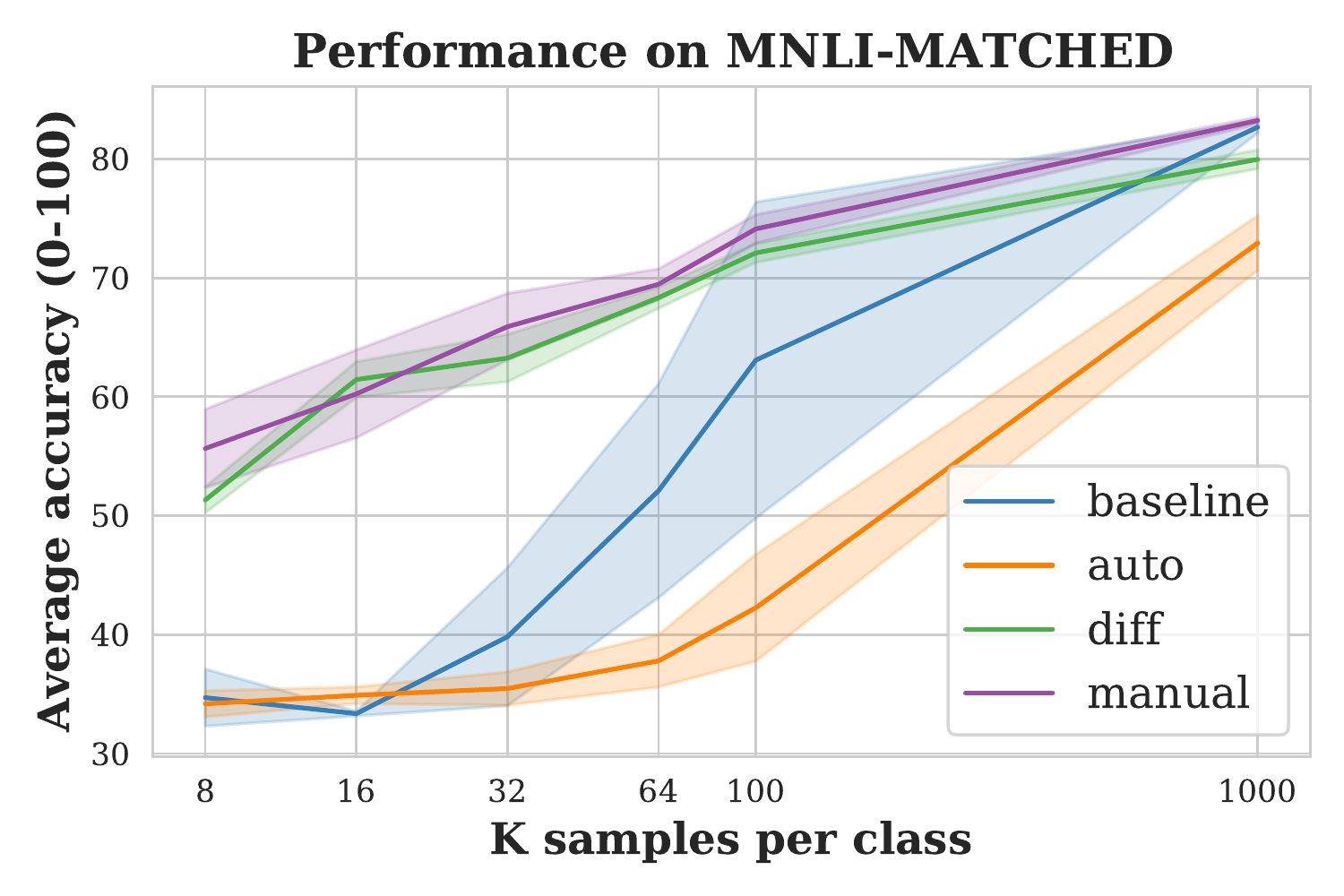}
    \caption{MNLI-MATCHED}
    \label{fig:mnli_matched_wider_k}
\end{subfigure}%
\begin{subfigure}{.33\textwidth}
  \centering
    \includegraphics[width=\hsize]{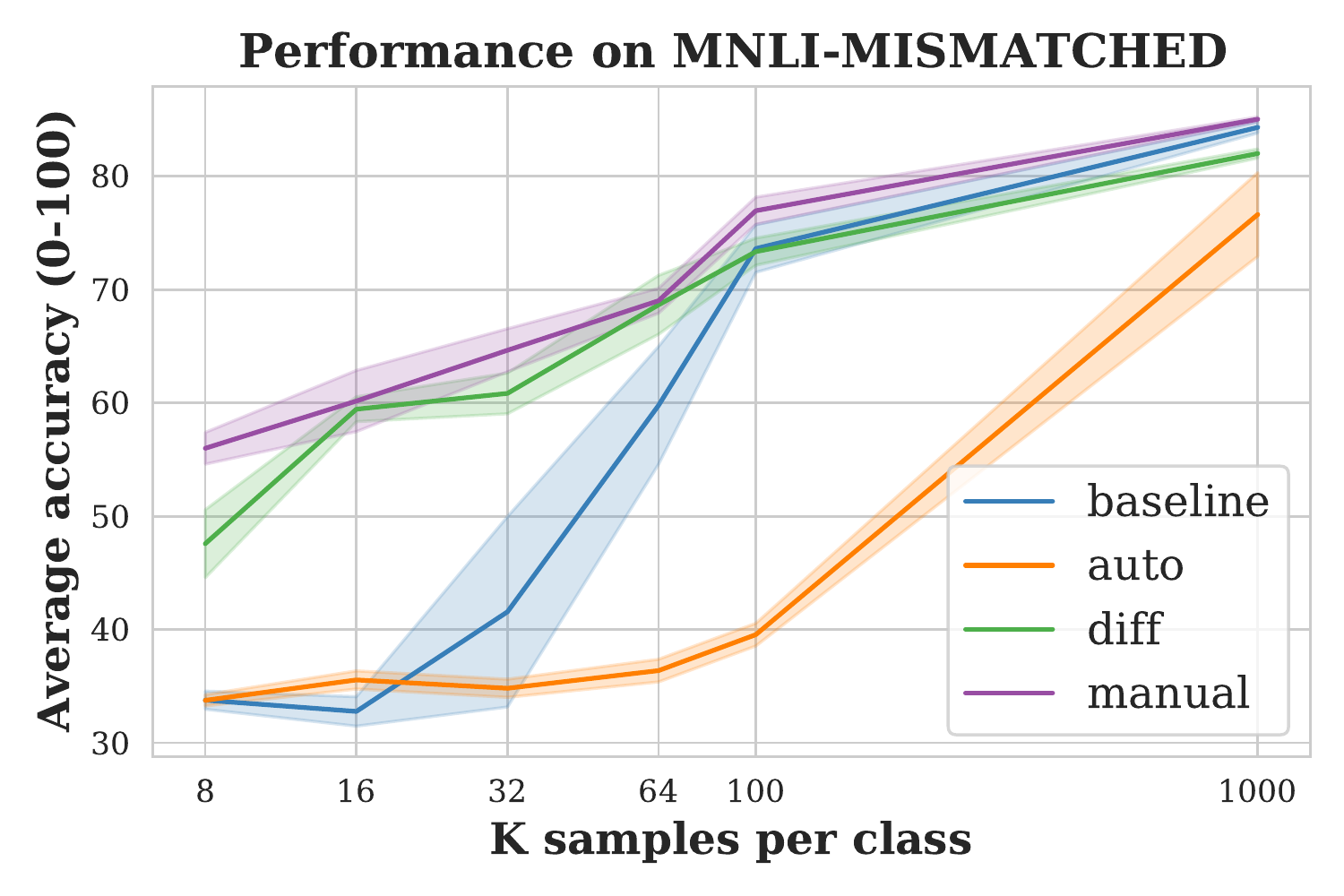}
    \caption{MNLI-MISMATCHED}
    \label{fig:mnli_mismatched_wider_k}
\end{subfigure}
\begin{subfigure}{.33\textwidth}
  \centering
    \includegraphics[width=\hsize]{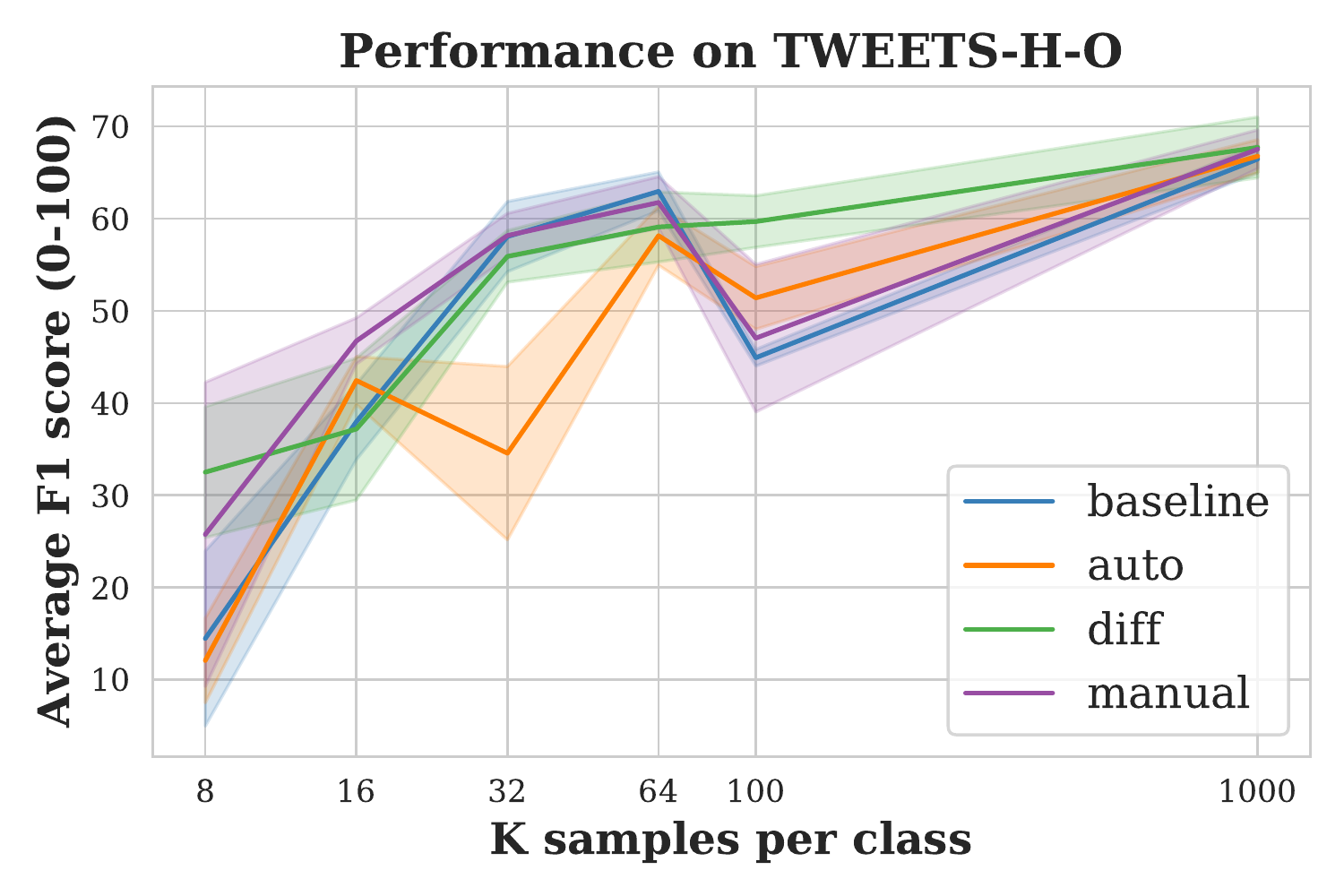}
    \caption{TWEETS-HATE-OFFENSIVE}
    \label{fig:tweets_wider_k}
\end{subfigure}
\vspace{-10pt}
\caption{The performance of prompting models on MNLI-MATCHED, MNLI-MISMATCHED \cite{gluedataset2018} and TWEETS-HATE-OFFENSIVE \cite{tweetsho2017} is shown for a wider range of $K$ values. The solid line plots the mean accuracy across five independent runs, and is bounded by one standard deviation on both sides.}
\label{fig:appendix_more_k}
\vspace{-5pt}
\end{figure*}
\section{Additional results for more K-shot experiments}
\label{sec:appendix:kshot}
In \Cref{fig:more_k} (\Cref{sec:eval:kshot}), we show the performance with more $K$ values for SST2, QNLI and ENRON-SPAM. Additional results in the same setup are shown in \Cref{fig:appendix_more_k}.

\end{document}